\documentclass[letterpaper, 10pt, journal, twoside]{IEEEtran}  % Comment this line out if you need a4paper

\IEEEoverridecommandlockouts                              % This command is only needed if 

\usepackage{graphics} % for pdf, bitmapped graphics files
\usepackage{epsfig} % for postscript graphics files
\usepackage{amsmath}
\usepackage{xcolor}
\usepackage{subfig}
\usepackage{graphicx}
\usepackage{amsfonts}
\usepackage{float}
\usepackage{algorithm}
\usepackage{algorithmicx}
\usepackage{algpseudocode}
\usepackage{multirow}
\usepackage[textwidth=1.5cm]{todonotes}
\usepackage{cite}
\usepackage{aecompl}
\usepackage{hyperref}
\usepackage{fancyhdr}

\title{
DROID: Minimizing the Reality Gap using Single-Shot Human Demonstration} 

\author{Ya-Yen Tsai$^{1}$, Hui Xu$^{2}$, Zihan Ding$^{3}$, Chong Zhang$^{3}$, Edward Johns$^{1}$, and Bidan Huang$^{3 \dag}$% <-this % stops a space
\vspace{-2.5mm}

\thanks{$\dag$ denotes the corresponding author.}% <-this % stops a space

\thanks{$^{1}$Y.-Y. Tsai is with the Hamlyn Centre for Robotic Surgery and E. Johns is with the Robot Learning Lab, Department of Computing, Imperial College London, SW7 2AZ, London, UK {\tt\footnotesize \{y.tsai17, e.johns\}@imperial.ac.uk}}
\thanks{$^{2}$H. Xu is with School of Computer Science and Engineering, University of Electronic Science and Technology of China {\tt\footnotesize hui\_xu@std.uestc.edu.cn}}
\thanks{$^{3}$Z. Ding, C. Zhang, and B. Huang,  are with Tencent Robotics X, China  {\tt\footnotesize \{zihan.ding18, chongzzhang, bidanhuang\}@tencent.com}}}

\begin{document}

\maketitle

%%%%%%%%%%%%%%%%%%%%%%%%%%%%%%%%%%%%%%%%%%%%%%%%%%%%%%%%%%%%%%%%%%%%%%%%%%%%%%%%
\begin{abstract}
Reinforcement learning (RL) has demonstrated great success in the past several years. However, most of the scenarios focus on simulated environments. One of the main challenges of transferring the policy learned in a simulated environment to real world, is the discrepancy between the dynamics of the two environments. In prior works, Domain Randomization (DR) has been used to address the reality gap for both robotic locomotion and manipulation tasks. In this paper, we propose Domain Randomization Optimization IDentification (DROID), a novel framework to exploit single-shot human demonstration for identifying the simulator's distribution of dynamics parameters, and apply it to training a policy on a door opening task. Our results show that the proposed framework can identify the difference in dynamics between the simulated and the real worlds, and thus improve policy transfer by optimizing the simulator's randomization ranges. We further illustrate that based on these same identified parameters, our method can generalize the learned policy to different but related tasks.
\end{abstract}
\section{INTRODUCTION}
% RL in robotics
Reinforcement Learning (RL) has been widely applied to decision making, control, and planning. In the field of robot learning, many works have adopted RL as the robot controlling policy, to improve its learning efficiency and performance~\cite{rusu2017sim, lin1993scaling, kober2013reinforcement}. Recent works have demonstrated that RL can be used to control a dexterous robotic hand or a robotic arm to solve tasks that require complicated manipulation skill, such as solving a Rubik's cube~\cite{andrychowicz2020learning, akkaya2019solving} or opening a door~\cite{urakami2019doorgym}. 

% Sim2real
RL uses self-exploration to find the optimal policy. Typically, this requires a very large amount of trial-and-error, which is time-consuming and can easily result in hardware damage if executed on the physical robot. A less costly and safer approach is learning the policy via simulation. However, the discrepancy between the real world and the simulation models could hinder the policy from directly being deployed in real world, especially when the task is contact-rich. In the literature, this \textit{sim-to-real} problem is referred to as the reality gap, which remains an open issue to date.

\textit{System Identification} (SI) and \textit{Domain Randomization} (DR) are the two common approaches to cross the dynamics reality gap. While SI tries to reduce the reality gap by identification of the real-world parameters, DR tries to increase robustness to the reality gap by training on randomized simulated environments. However, these methods still often struggle to accurately obtain real-world parameters or choose randomization ranges without any real data~\cite{valassakis2020crossing}. This can result in learning a biased policy, or a policy which fails to converge.

\begin{figure}
    \centering
    \includegraphics[width=8cm]{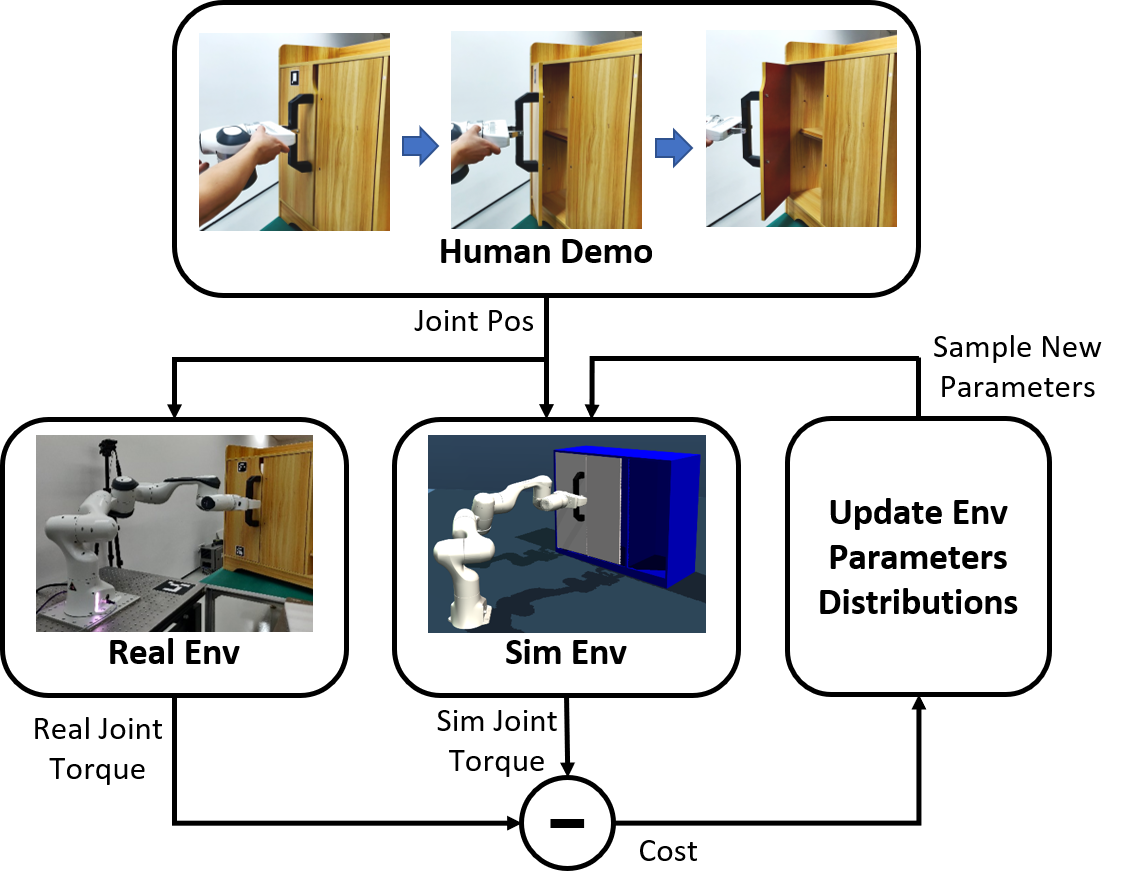}
    \caption{This figure presents the overview of the proposed framework, DROID, used to minimize the reality gap. }
    \vspace{-1.5mm}
    \label{fig:overview}
\end{figure}

To address this, we propose a novel framework, Domain Randomization Optimization IDentification (DROID), to automatically optimize the environment parameter distribution, with a combination of DR and SI approaches. Rather than determining the DR range using intuition or tedious tuning, we exploit human demonstrations and attempt to align simulated trajectories with these real-world trajectories. Through this, we identify the simulator's optimal parameter range as a statistical model, which can then be sampled from during training with RL. An overview of this method is shown in Fig.~\ref{fig:overview}.

With DROID, the learned policy can be transferred to the real world directly, thus learning efficiency is significantly improved and unsafe interactions with the environment is avoided. Our experimental results show that after parameter optimization and identification, a much higher success rate can be achieved with DROID on a real-world door opening task, compared with standard DR or SI. We also show how the learned dynamics can be directly used to train policies for different but related tasks.

\section{RELATED WORK}
% sim2real: what is reality gap?
RL's applications in robotics suffer from expensive training data collection in real robots~\cite{levine2018learning}. By offering cheaper and safer data collection environment, simulation has gained huge popularity for RL training~\cite{brockman2016openai, andrychowicz2020learning, tsai2020constrained, garcia-hernando2020physics}. However, the issue of the reality gap still remains one of the main problems hindering the policy learned in simulation from transferring well to the real world. To bridge the reality gap, two common strategies have been studied in prior works: SI, and DR~\cite{valassakis2020crossing}.

SI has been a popular approach for sim-to-real transfer~\cite{yu2017preparing, allevato2020tunenet, jeong2019modelling, liang2020learning, kaspar2020sim2real, schafroth2010modeling}. It focuses on finding the exact model of the real world, so the physical behaviors match between the real and simulated system. This could be done by constructing the simulation through direct measurement of the environment parameters in the real world~\cite{tan2016} or by collecting real world data for optimizing the simulated model parameters~\cite{farchy2013humanoid,yu2019sim,tobin2017domain}. However, correctly identifying the system's parameters is challenging. Many parameters cannot be explicitly measured or can involve presence of noises, especially for dynamics related ones, like friction, stiffness, and damping. In addition, the system parameters could be of high-dimensional and entangled. This further increases the difficulties in achieving accurate and precise SI results~\cite{zhu2018efficient, fazeli2018identifiability, valassakis2020crossing}. As a consequence, SI usually requires expertise of the system to handcraft the model.

% DR
Rather than identifying the environment parameters, DR creates multiple simulated environments by randomizing the system parameters within given ranges during policy training. This improves the policy's generalizability and robustness against the reality gap. Recently, it has achieved significant progress for sim-to-real transfer in robotics~\cite{valassakis2020crossing,peng2018sim, james2017transferring,chebotar2019closing,ramos2019bayessim, andrychowicz2020learning, akkaya2019solving, tobin2017domain,urakami2019doorgym,alghonaim2020benchmarking}. Unlike SI, DR achieves better sim-to-real performance by covering a greater range of parameters distribution in the simulation containing the real values during RL training. Prior works optimize system parameters with simulated and real trajectories collected using hand-designed policies~\cite{andrychowicz2020learning}; or automatically adjust the boundaries of uniform randomization distributions according to model performances\cite{akkaya2019solving}. While effective, these works suffer from a common drawback. Existing works~\cite{valassakis2020crossing} have demonstrated its demanding engineering efforts in adjusting the randomization ranges, which is difficult and not intuitive. Hand tuning the parameter ranges could easily cause overestimated values and lead to training RL in an invalid environment and learning a suboptimal policy. How to quickly choose the randomization ranges for different parameters and achieve effective policy generalization still remains a challenge.

Besides the two main approaches, a variant of DR, Adaptive DR, has also been studied. It was proposed to optimize the parameter distributions and minimize the chance of training RL in invalid environments. Prior works use techniques such as approximate Bayesian computation~\cite{beaumont2002approximate}, Bayesian Optimization, or a relative entropy policy search~\cite{peters2010relative} to estimate or optimize distributions of system parameters~\cite{ramos2019bayessim,muratore2020bayesian,chebotar2019closing}. The proposed framework draws some similarities to these works in avoiding overestimation of parameter distribution, but focuses on more contact-rich task which involves determining complicated and entangled dynamics related parameters and optimizing the distribution through a more efficient and safer approach, i.e. human demonstration. In addition, we optimize the randomization distributions with respect to trajectories containing not only the observed positions and/or velocities, but also the proprioceptive torques on joints of the robot arm. Experiments were conducted on a contact-rich task, door opening, with DROID due to its complex dynamics of the robot joints, the door hinge and the contacts between the gripper and the handle.

The rest of the paper is organized as follows: The methodologies is presented in Section III, followed by  the experimental setup, results and the discussion section in IV. Finally, the conclusions and the future works are presented in Section V.

% contributions

\section{METHODOLOGY}
\label{sec:method}
% Overview
Learning in a simulated environment is convenient, but transferring the learning results to the real environment requires an accurate model of that environment. Simulation typically uses mathematical models to compute the interaction force and torque between objects. These models rely on pre-defined dynamics-related parameters such as friction, stiffness and damping. Unlike kinematics-related parameters, many of them are not easily accessible and hence are often difficult to measure and identify. Therefore, tasks that involve these parameters experience difficulty in sim-to-real transfer. To this end, we propose a framework DROID that evaluates these parameters through human demonstration. Building on the concept of interaction force, we implicitly perceive dynamics information of the real-world system from the feedback of the robot and use this information to determine the distribution of the parameters in the simulation. This gives us a reasonable set of parameters for the domain randomization in RL and hence results in a successful policy transfer.

DROID is composed of three phases: the human demonstration (Section~\ref{sec:sec:data}), the parameter identification and optimization (Section~\ref{sec:sec:para}), and the policy learning with optimized DR  (Section~\ref{sec:sec:policy}). In the first phase, the human demonstrates a contact task in the real world. The robot clones the human behaviors multiple times and records the data. In the second phase, robot in the simulator repeats the same behaviors and records a same set of data.
The data from the real system and the simulator is then used for identifying the distribution of the task relevant parameters.  
Through an iterative approach, we can gradually update the parameter distribution to minimize the differences between the two perceived feedback until the obtained simulated environments can better reflect to the real world. In the final phase, policy is trained with DR based on the parameter distribution optimized. The resulting policy can be transferred to the real with good performance for the previous optimization steps. Note that this study focuses on minimizing the reality gap and the human demonstrations only serve for the purpose of identifying the task relevant parameter distribution. 
Learning the task from human demonstration is out of the scope of this paper and in the third phase we learn the policy from scratch without human demonstrations. In the following sections, we will go into more details on how each part is implemented. 

% Human Demonstrations and data collection
\subsection{Single-Shot Human Demonstration}
\label{sec:sec:data}
In this first phase, our aim is to collect the data reflecting dynamics relevant to the task. To this end, human demonstrates the task once in the real world to provide a robot motion trajectory, $\boldsymbol{q_{d}}$, through kinesthetic guidance. This demonstration is safe to be executed by the robot and allows the robot to repeat it multiple times automatically. By repeating the $\boldsymbol{q_{d}}$ to interact with the environment, the dynamics information can be perceived by the torque sensor feedback $\boldsymbol{\tau_r}$ of the robot. Such feedback is later used as the reference to identify and update the parameter distribution in the simulation. Different from the approaches that makes the robot to randomly interact with the environment, in DROID we rely on the human to provide a trajectory that is safe for the robot to identify the system dynamics. This only requires a single-shot demonstration. We focus on the task relevant parameter distributions and limit the random exploration of the robot in the real world. This minimizes the risk and save the time in the real robot experiment.

% Parameter Opt and ID
\subsection{Parameter Optimization and Identification}
\label{sec:sec:para}
The goal in this phase is to correctly identify the parameter distribution that minimizes the discrepancy of the simulated and the real system. Rather than finding the specific value for each parameter, we determine the distributions of them for the presence of noises and uncertainties in the real world. We can then train the RL to find a policy that works within this distribution, which is the key problem solved in DROID. 
Falsely defined distribution will lead to failure in sim-to-real transfer, and the policy trained under an unreasonable DR can suffer bad performance. 
We model this distribution as a multivariate normal distribution ${\Phi}(\boldsymbol{\mu},\boldsymbol{\Sigma})$, where $\boldsymbol{\mu}$ and the $\boldsymbol{\Sigma}$ is the mean and covariance, and system parameters $\boldsymbol\phi$ is randomly sampled from ${\Phi}$. 

% details of the proposed algorithm
During the human demonstration phase, data has been collected from the real system. Taking the identical steps, we can program the robot to repeat the same task in the simulation and hence obtain the torque sensor feedback $\boldsymbol{\tau_s}$. As the real world parameter value $\boldsymbol\phi^{\prime}$ is not easily accessible, we align the simulation and the real environment by aligning the robot behaviors in them. Here, we define the ``behavior'' as the robot action and perception pairs, i.e. the motion trajectory $\boldsymbol{q_{d}}$ and the torque sensing $\boldsymbol{\tau_s}$. Changing $\boldsymbol\phi$ changes the dynamics of the simulation and hence changes the robot behavior. Sampling different $\boldsymbol\phi$ from its distribution ${\Phi}$, we observe the different behaviors under different environments in the simulator and identify the ones that are most similar to the real world behavior. We hence update ${\Phi}$ based on the robot behaviors. 

We formulate this as a distribution optimization problem and update $\Phi$ iteratively based on the Covariance Matrix Adaptation Evolution Strategy (CMA-ES) approach~\cite{hansen2006cma} with the following objective function:
\begin{equation}
    \mathcal{J}(\boldsymbol{\phi}) = \dfrac{1}{N}\sum_{n=1}^N (\|\boldsymbol{\tau_s}(\boldsymbol{\phi}) - \boldsymbol{\tau_r}^n(\boldsymbol{\phi}^\prime)\| + c\beta), \boldsymbol{\phi}\sim\Phi
    \label{eqn:cost}
\end{equation}

where $N$, $\beta$, $c$ is the total number of trajectories from the real robot, a penalty for failure of the task and the factor for the penalty, respectively.

 The process is iterated from the initial guess, $\Phi_{init}=\mathcal{N}(\boldsymbol\mu_{init}, \boldsymbol{\Sigma}_{init})$, until convergence. Note that we append $c\beta$ at the end of the fitness function to penalize the situation where the robot fails to grip the door knob during door opening process in the simulation. This occasionally happens when for example the friction coefficients of the fingers become to low or when the joint damping is set to be invalid. The overview of the described algorithm is summarized in Algorithm~\ref{alg:opt}. $M$ is the total number of samples from the current distribution $\Phi$.
For each iteration, the robot performs the task in the simulator under the environment parameter $\boldsymbol{\phi}$ sampled from the distributions ${\Phi}$. The cost $\mathcal{J}(\boldsymbol{\phi})$ is therefore evaluated with the resulting $\boldsymbol{\tau_{s}}$ in Eqn.~\ref{eqn:cost}. Note that in total $N$ real robot trajectories are used to evaluate the cost and $\mathcal{J}(\boldsymbol{\phi})$ is the average value. After $M$ iterations, the CMA-ES updates ${\Phi}$ from the $x$ best candidates which associated with the lowest costs $\mathcal{J}(\boldsymbol{\phi})$. The update process of ${\Phi}$ (\emph{i.e.}, $\boldsymbol\mu$ and $\boldsymbol\Sigma$) and hyper-parameters of CMA-ES following the standard procedure. This repeats until ${\Phi}$ converges and we achieve the optimized $\Phi^*$

\begin{algorithm}
\caption{Optimizing parameter distribution}
\begin{algorithmic}[1]
\State Initialize hyper-parameters of CMA-ES
\State Initialize $\Phi$ with $\mathcal{N}(\boldsymbol\mu_{init}, \boldsymbol{\Sigma}_{init})$
\While{not converged}
    \For {$m = 1:M$}
    \State Sample $\boldsymbol{\phi_m}$ from $\Phi$
    \State Robot perform task in simulation with $\boldsymbol{\phi_m}$ 
    \State Collect $\boldsymbol{\tau_s}(\boldsymbol{\phi_m})$
        \State Calculate $\mathcal{J}(\boldsymbol{\phi_m})$ in Eqn.\ref{eqn:cost} with $N$ real trajectories $\{\boldsymbol{\tau_r}^1(\phi^\prime),..., \boldsymbol{\tau_r}^N(\phi^\prime)\}$
    \EndFor
    \State Select $x$ best $\boldsymbol{\phi}$ from $\{\boldsymbol{\phi_1}, ..., \boldsymbol{\phi_M}\}$ by $\min \mathcal{J}(\boldsymbol{\phi})$
    \State Update $\Phi=\mathcal{N}\left(\boldsymbol\mu,\boldsymbol\Sigma\right)$ with the selected $\boldsymbol{\phi}$ set
    \State Update hyper-parameters of CMA-ES
\EndWhile
\State\Return optimized $\Phi^*$
\end{algorithmic}
\label{alg:opt}
\end{algorithm}

\begin{figure*}[ht!]
    \centering
    \subfloat[]
    {\includegraphics[width=0.24\textwidth]{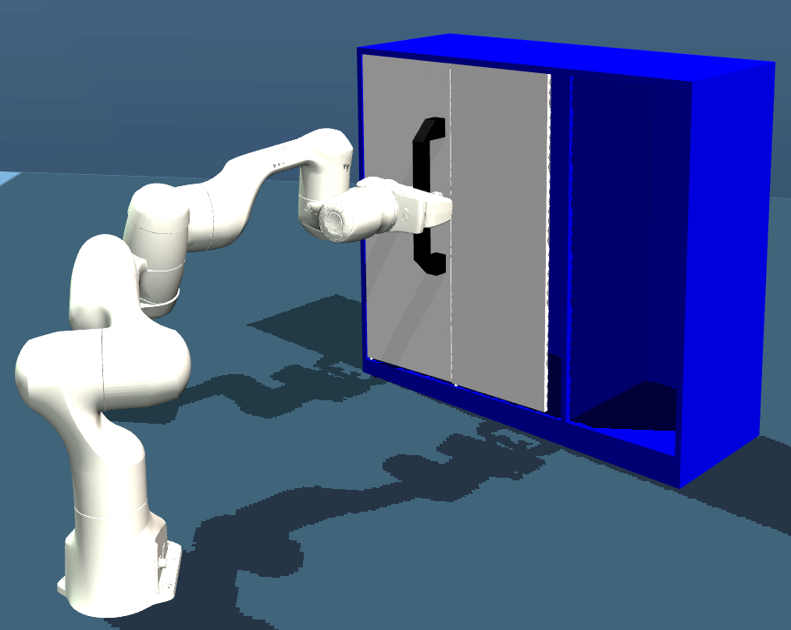}}
    \hspace{1mm}
    \subfloat[]
    {\includegraphics[width=0.24\textwidth]{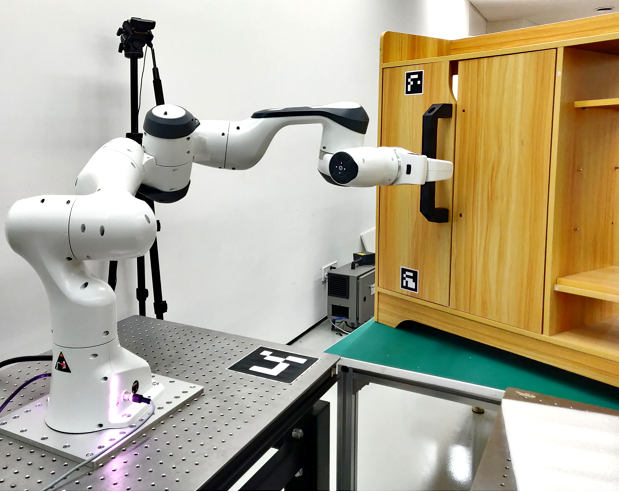}}
    \hspace{1mm}
    \subfloat[]
    {\includegraphics[width=0.24\textwidth]{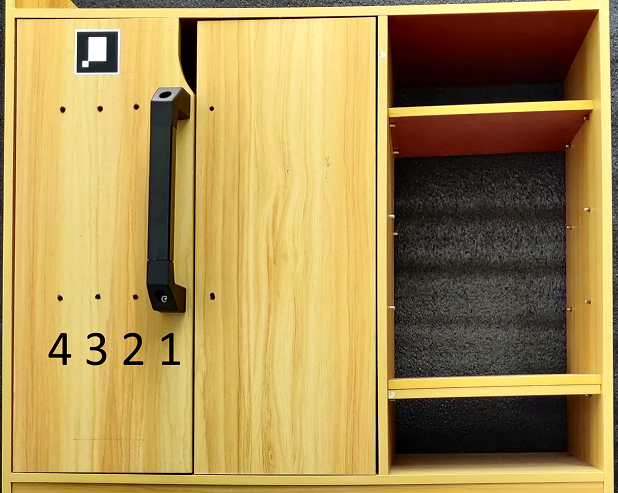}}
    \hspace{1mm}
    \subfloat[]
    {\includegraphics[width=0.24\textwidth]{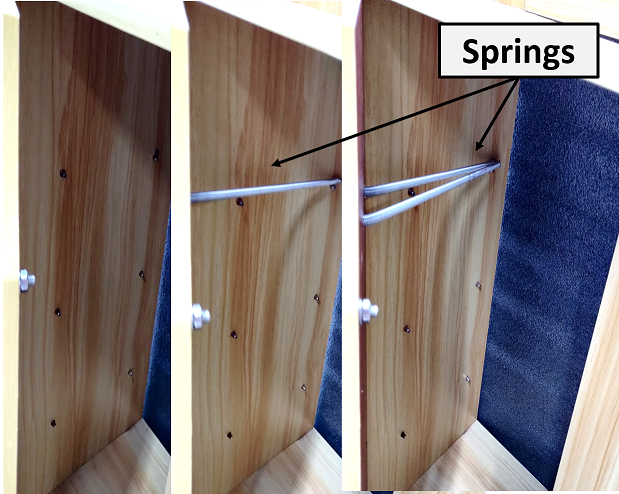}}
    \caption{(a) and (b) shows the hardware setup in the real world and simulation. The AruCo markers attached to the table and the cabinet are used for the tracking purpose. (c) shows four different locations of the door knob, each separated by 5\emph{cm} along the lever arm of the door. (d) shows three variants of the door. From left to right are the original door, door with one spring and door with two springs attached to the hinge. The springs were used to emulate doors with different dynamics.}
    \label{fig:setup}
    \vspace{-3mm}
\end{figure*}

% Policy Learning (Reinforcement Learning/Transfer Learning)
\subsection{Policy Learning}
\label{sec:sec:policy}
In this paper, we adopt a reinforcement learning framework to learn an optimal control policy $\pi_\theta$, parameterized by $\theta$, through policy gradient. This is formulated as a Markov decision process (MDP) which is defined by the tuple $(\mathcal{S}, \mathcal{A}, \mathcal{P}, r, \rho_0, \gamma )$, where $\mathcal{S}$ is the state space, $\mathcal{A}$ is the action space, $\mathcal{P} : p
\left( s _ { t + 1 } | s _ { t } , a _ { t } \right)$ is a
distribution of state transition,
$\mathcal{R}: \mathcal{S} \times \mathcal{A} \rightarrow \mathbb{R}$
is the reward function, $\rho_0$ the initial state distribution and
$\gamma \in ( 0,1 )$ is the discount factor.
The optimal policy $\pi^*_\theta$ aims to maximize the cumulative reward over episodes:
\vspace{-1mm}
\begin{equation}
    \pi^*_\theta = \mathop{\arg\max}_{\pi_\theta} \mathbb{E}_{\pi_{\theta}}\left[\sum_{t}
    r\left(s_{t}\right)\right].
    \label{eqn:optimal_pi}
\end{equation}

Proximal Policy Optimization (PPO) \cite{schulman2017proximal} is deployed for the robot learning purpose. It updates the policy by using the surrogate objective: 
\vspace{-1mm}
\begin{equation}
    L(\theta) \\
    =E_t[min(r_t(\theta)\hat{A}_t, clip(r_t(\theta), 1-\epsilon , 1 + \epsilon)\hat{A}_t)],
    \label{eqn:ppo}
\end{equation}
where, $\hat{A}_t$ is the estimate of the advantage function at timestep $t$, and $r_t(\theta) $ denotes the ratio between the current policy and the previous policy. The clipped term keeps the ratio inside the interval $[1-\epsilon, 1+\epsilon]$, the minimum of the clipped and unclipped term is used for the expectation. This provides a lower bound, or a pessimistic bound, on the unclipped term.

As the model of the real world environment is described by $\Phi^*$, sufficient simulation environments are hence required to be sampled from the distribution in order to reflect to the reality. Therefore, the RL agent is trained on multiple simulated environments sampled from $\Phi^*$. By doing so, we hope to maximize the similarity in state transition between the simulated and the real environments and enhance the transferrability of the learned policy to the real world application. 

% reward function
The structure of the reward function follows closely to the one defined in \cite{urakami2019doorgym} and is presented as follows:

\vspace{-5mm}

\begin{equation}
 r =
\begin{cases}
    \omega_1 \cdot r_{door} + \omega_2 \cdot r_{ori}\\ \quad+ \omega_3 \cdot r_{dist} + \omega_4 \cdot r_{log\_dist} + \omega_5 \cdot r_{slip} , \text{ if } \lambda <30^\circ,\\ \\
    \omega_1 \cdot r_{door} + \omega_2 \cdot r_{ori} + \omega_5 \cdot r_{slip}, \text{ otherwise} 
\end{cases}
\end{equation}
where $\lambda $ is the hinge angle of the door, ranging from 0$^\circ$ (closed door) to 90$^\circ$ (completely opened door).
The reward function has a total five terms.  The $r_{door}$ rewards the action that results in the increase to the door hinge. The $r_{ori}$ rewards the relative orientation between the door knob and the gripper. The higher reward is given if the relative orientation is closer to orthogonal. The $r_{dist}$ and $r_{log\_dist}$ are associated with the relative displacement between the door knob and the gripper. 
The higher reward is given to the short displacement. One major difference between our door opening strategy and the DoorGym's, lies in that they use hooks to open the door while our robot is trained to firmly grip knob handle during the entire task. This significantly increases the complexity of dynamics involved. To this end, we added a penalty term $r_{slip}$ in the reward function to prevent the fingers from slipping. $\omega_1, \omega_2, \omega_3, \omega_4$, and $\omega_5$ are five coefficients for normalizing each reward terms. 
\section{EXPERIMENTS AND RESULTS}
We evaluate our work through a door opening task for its complexity in the dynamics involves. The main focus of our experiments is to access how well the RL policy learned with DROID can be transferred to the robot in the real world. This evaluation consists of three experiments. 

The first experiment validated that DROID can identify the parameter distributions for environments and with different dynamics. In the second part, we applied one of the optimized distributions to DR and train a policy for door opening. The evaluation on the performance of this policy was done by comparing it's robustness and effectiveness in sim-to-real transfer with other approaches (standard DR and without DR). Finally, we tested the generalizability of the RL policy learned from the optimized parameter distributions.

\subsection{Experimental Setup}
\label{sec:sec:setup}

The experiment consisted of a real and a simulated system. In the real system, a cabinet door, a camera and a 7-DoF Franka Emika robot arm were used as illustrated in Fig.~\ref{fig:setup}(a)(b). The Franka robot was equipped with 7-DoF joint torque sensors allowing it to record the torque feedback while interacting with the environment. A two fingers gripper was mounted at the end effector for grasping and manipulation purpose. The cabinet was the target of interest in which we attempted to identify its dynamics parameter distributions. The fixed camera provided the relative pose information between the robot and the door as well as the door angle information via tracking the visual markers attached to the door and the optical table during the experiment. 

For the simulation part, we deployed the MuJoCo platform~\cite{todorov2012mujoco} to perform the RL training. The simulated environment was defined by a set of parameters including kinematics tree, and many dynamics such as mass, damping, friction etc. Many kinematics such as relative poses and the geometric dimensions and the robot related dynamics such as mass and inertia were either provided officially or can be directly measured and hence were not within our focus. Inertia of the door was also not considered as it can be estimated once we obtained the CAD model. Our main interest was to identify the parameter distributions of the dynamics of the robot and the door that were not provided and difficult to measured in the real world. These parameters included mass, joint friction loss, and joint damping and the sliding and torsional frictions.

\subsection{Parameter distribution identification}
% CMA-ES
In the first experiment, we investigate the feasibility of DROID in the distributions identification. This verifies whether DROID can find the distribution of parameters that correctly reflects the interaction behavior encountered in the real-world door opening  task. 

For the given cabinet door (Fig.~\ref{fig:setup}), the real robot followed the human demonstration to obtain ten sets of $\boldsymbol{\tau_r}$ that reflected the dynamics of the interactions with the door. We first made an initial guess of the $\Phi_{init}$ for the parameters of interest. The estimation was made by referencing Franka's officially provided values and DoorGym's parameters~\cite{urakami2019doorgym} with the exceptions being the door and knob masses which were directly measured. At the first iteration, 30 simulated environments were sampled from the initial distributions. Parameters sampled with negative values were omitted and resampled. With each environment, the robot cloned the human demonstration by following $\boldsymbol{q_{d}}$ to obtain $\boldsymbol{\tau_s}$. The associated cost for each trail was calculated using Eqn.~\ref{eqn:cost}. The higher the discrepancy among the torques, the higher the cost would be gained. For the simulation that failed to successfully open the door due to factors like grasp slipping, an extra penalty of 10 was added to the cost. CMA-ES algorithm took the top five best candidates of $\boldsymbol{\phi}$ to update the means and the covariances and hence the $\Phi$. In a new iteration, the updated $\Phi$ were used to generate another 30 new simulations and this process was iterated until convergence. 

\begin{figure}[]
    \centering
        \subfloat[]
    {\includegraphics[width=0.4\textwidth]{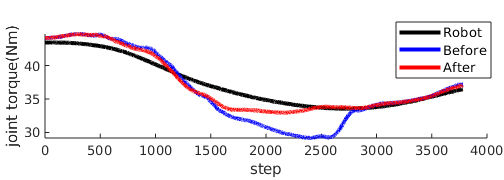}}
    \vspace{1mm}
    \subfloat[]
    {\includegraphics[width=0.4\textwidth]{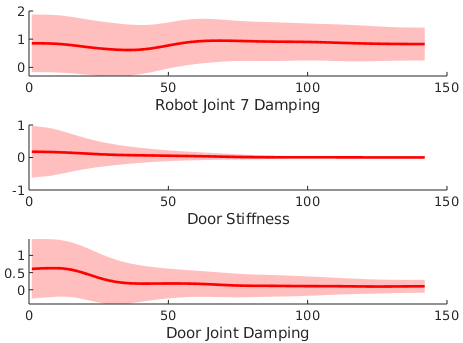}}
    \vspace{1mm}
    \subfloat[]
    {\includegraphics[width=0.41\textwidth]{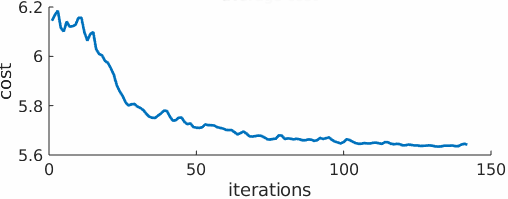}}
    
    \caption{(a) shows the comparison between a robot joint torque trajectory in real world and the joint torque trajectories before and after the optimization and identification for parameter distribution in simulation. (b) shows the changes in means (the solid lines) and the variances (the shaded areas) of the three parameters over iterations during optimization. The red line represent the mean and the shaded region represent the variance. (c) shows the associated cost computed using Eqn.~\ref{eqn:cost} over iterations. }
    \label{fig:traj_comp}
    \vspace{-5mm}
\end{figure}

With the above steps, we have estimated the parameter distribution for our cabinet door (Tab.~\ref{tab:init_final_param}). These parameters reflect the dynamics of the robot arm, of the door hinge and the contacts. Fig.~\ref{fig:traj_comp} illustrates the optimization process and results. Fig.~\ref{fig:traj_comp}(a) shows the joint torque trajectories of the robot obtained in the simulation before and after the optimization, and the joint torque trajectory obtained in the real world. This only displays one of the joint for illustration. As it can be seen, differences between the red (after optimization) and the black (real robot) lines are much smaller than the differences between the blue (before optimization) and the black. This suggests that the proposed approach can indeed minimize the reality gap. The same conclusion can be drawn from Fig.~\ref{fig:traj_comp}(c) plotting the cost against the iteration. Among the 30 simulations the average cost gradually converged to a lower value. This indicates that the simulations sampled from the optimized distribution lead to smaller reality gap than those sampled from the initialed distributions. The means and the variances of three parameters at each iteration are shown in Fig.~\ref{fig:traj_comp}(c) and these show the distribution of these parameters converged to fixed ranges. The quantitative results comparing the unoptimized and optimized distributions are summarized in Tab.~\ref{tab:init_final_param}. We have applied this result to the sim-to-real transfer experiment, which is detailed in the next section. 

\begin{figure}
    \centering
    \subfloat[]
    {\includegraphics[width=0.17\textwidth]{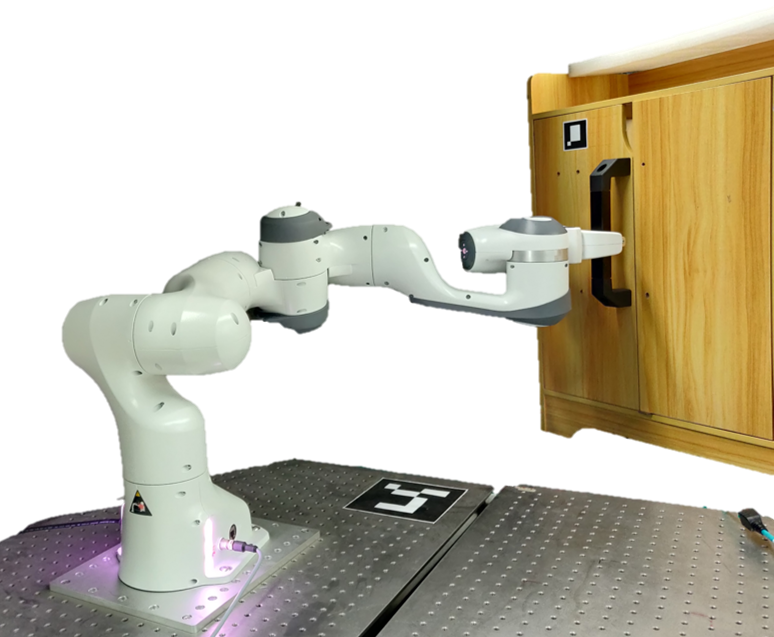}}
    \hspace{4mm}
    \subfloat[]
    {\includegraphics[width=0.17\textwidth]{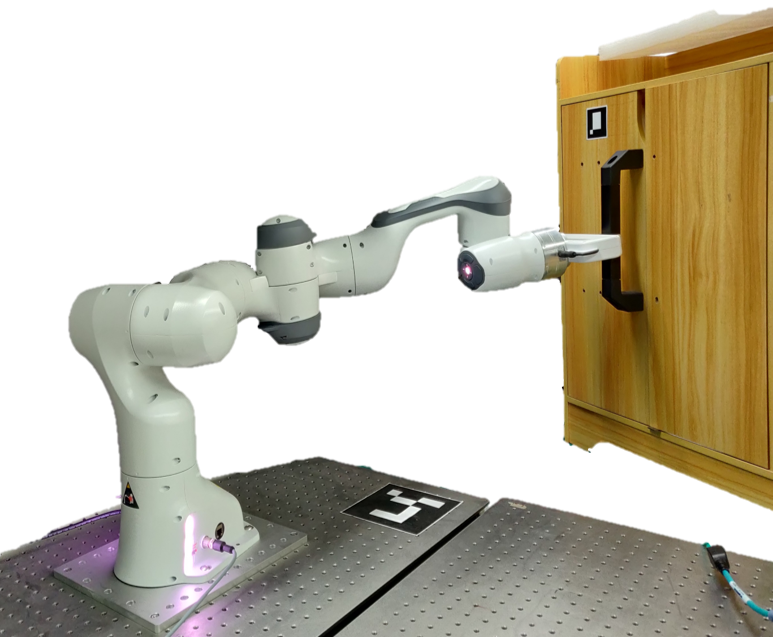}}

    \caption{Two different initial robot poses used in the human demonstrations. (a) is the initial robot pose used to estimate parameter distribution in Tab.~\ref{tab:init_final_param}. (b) is another initial robot pose used to validate the possibility of a bias in parameter estimation.}
    \vspace{-3mm}
    \label{fig:robot_pose}
\end{figure}

\begin{figure}
    \includegraphics[width=0.49\textwidth]{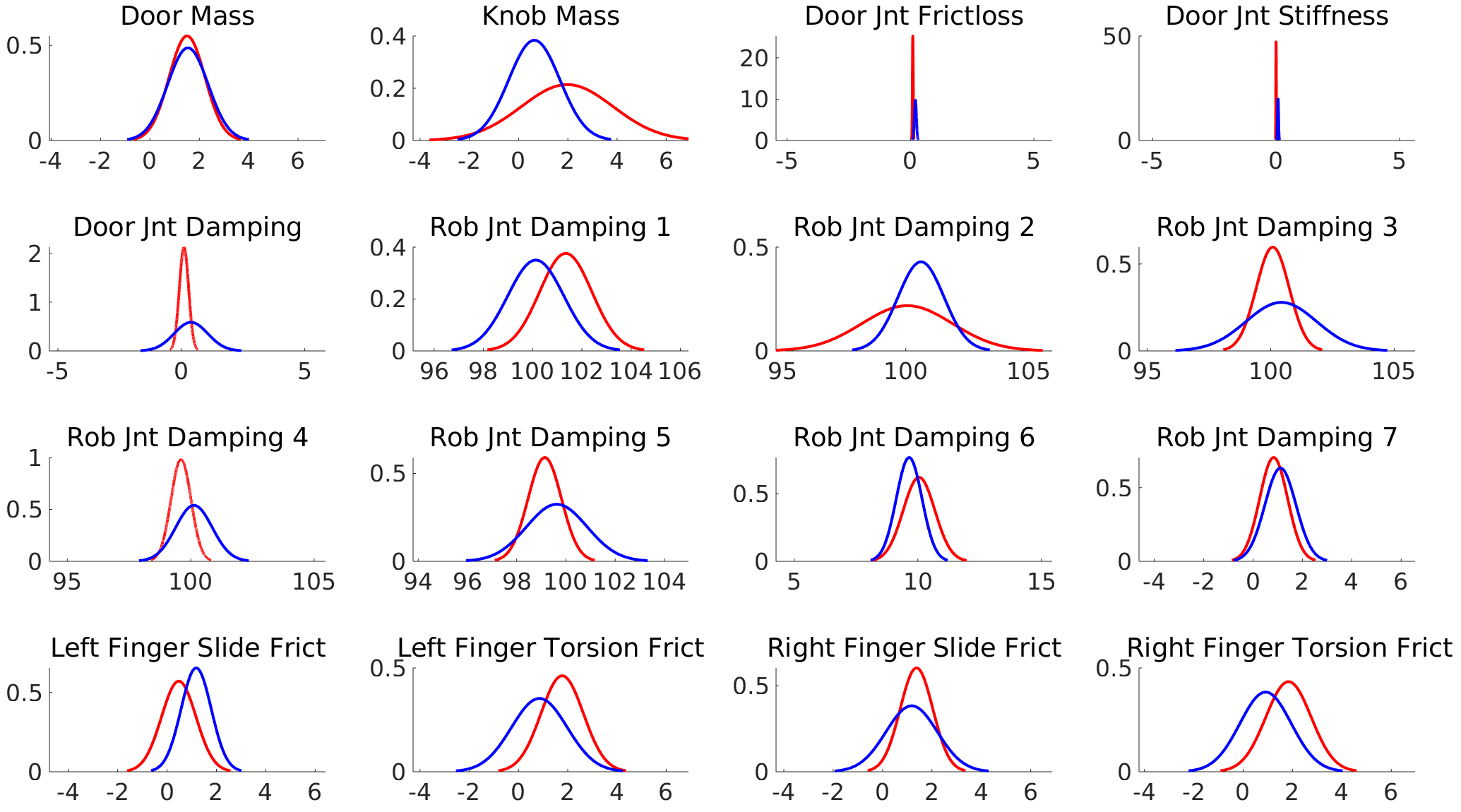}
    \caption{This compares the parameter distributions estimated using two different initial poses and human demonstrations. Red and blue curves represent the parameter distribution estimated using robot pose (a) and (b) shown in Fig.~\ref{fig:robot_pose}.}
    \vspace{-5mm}
    \label{fig:distribution}
\end{figure}

Furthermore, a validation was carried to determine the possibility variations in estimating the parameter distribution using different single human demonstration. Different human demonstration was provided through an alternative robot pose as illustrated in Fig.~\ref{fig:robot_pose} for validation. Note that due to robot workspace limitation, only the presented two poses could be applied to interact with and open the door. The parameter distribution identification process was repeated using this pose. The experimental result comparing the estimated parameter distribution using the two human demonstrations is shown in Fig.~\ref{fig:distribution}. We verified that despite for the minor variations, the majority of parameter distributions estimated using the alternative pose still converged to similar distributions as stated in Tab.~\ref{tab:init_final_param} and single human demonstration was adequate for parameter estimation.

In order to further verify the approach, we have conducted two sets of controlled experiments by changing the dynamics of the real door. 

The original door was modified to include additional springs to change its dynamics. Fig.~\ref{fig:setup}(d) illustrates the three variants of the door, without any spring, with one spring and with two springs attached to the hinge. These springs were identical and share similar dynamics. 

The optimization results for different door dynamics are also summarized in Tab.~\ref{tab:init_final_param}. It can be seen from the parameter values that the robot dynamics is not much affected but the environment dynamics varies a lots. This is due to the fact that we only changed the door dynamics and the algorithm is able to identify this change correctly. Doors equipped with different amount of springs would expect to have different distributions for the door joint frictionloss, stiffness and damping. For the frictionloss, door with 2 springs seem to have higher friction loss than the other two scenarios. The joint stiffness increases as the number of spring equipped increase, which is reasonable. Doors with springs tend to have higher joint damping values than the door without the spring as expected. The joint dampings associated to the robot falls into the similar values. This shows the proposed framework have consistant results. The masses and the frictions vary across different scenarios, and this may be attribute to the robot which needs higher friction in the gripper in order to sustain the sufficient force to grip the door knob in the simulation when more spring is loaded to the door. 

Based on the above evaluation results, the proposed DROID framework has demonstrated to be feasible in determining the parameter distribution of the real world dynamics. In theory, the reality gap is smaller after optimization. The framework has also shown to be realizable to identify task with different dynamics. Most outcomes have appeared to be reasonable with a few exceptions which may be attribute to factors such as imperfect demonstration resulting in loss of grip when door becomes stiffer or noises present in joint torque sensors in the real world. We have used the optimized result in sim-to-real practice and detail it in the Section~\ref{sec:sec:DR}.  

{\renewcommand{\arraystretch}{1.5}
\begin{table*}[]
\centering
\captionof{table}{This compares the initial and final parameters distribution after optimization and identification. The distribution is defined by $\mu$ and $\sigma$. Additionally, this also compares the optimized distributions for doors equipped with one or two springs.}
\label{tab:init_final_param}
\begin{tabular}{|c|c|c|c|c|c|c|c|c|}
\hline
& \multicolumn{2}{c|}{\textbf{DoorGym}}                                                                                      & \multicolumn{2}{c|}{\begin{tabular}[c]{@{}l@{}}\textbf{Door without spring}\end{tabular}}                                                   & \multicolumn{2}{c|}{\begin{tabular}[c]{@{}l@{}}\textbf{Door with $1$ spring}\end{tabular}}                                                & \multicolumn{2}{c|}{\begin{tabular}[c]{@{}l@{}}\textbf{Door with $2$ springs}\end{tabular}}   
\\ \hline
& $\mu_{init}$ & $diag(\Sigma_{init})$   & $\mu_{opt}$  & $diag(\Sigma_{opt})$  
& $\mu_{opt}$  & $diag(\Sigma_{opt})$    & $\mu_{opt}$  & $diag(\Sigma_{opt})$      \\    \hline
\multicolumn{9}{|c|}{\textbf{Door Properties}}  \\ \hline
Door Mass(kg)   & $1.144$    & 0.5    & 1.50   & 0.73    & 0.92  & 0.52   & 2.54   & 1.23      \\ \hline
Knob Mass(kg)   & 0.199      & 0.1    & 1.98   & 1.86    & 2.31  & 0.42   & 3.99   & 1.29      \\ \hline
\begin{tabular}[c]{@{}c@{}}Friction Loss\end{tabular}       
& 0.05      & 0.025     & 0.10   & 0.02   & 0.0     & 0.0     & 0.56   & 0.05              \\ \hline
\begin{tabular}[c]{@{}c@{}}Joint Stiffness\end{tabular}           
& 0.01      & 0.005     & 0.002   & 0.008  & 1.06    & 0.0     & 1.24  & 0.06             \\ \hline
\begin{tabular}[c]{@{}c@{}}Joint Damping\end{tabular}              
& 2.0        & 1.0     & 0.10    & 0.19    & 0.71    & 0.01    & 0.6    & 0.18              \\ \hline
\multicolumn{9}{|c|}{\textbf{Robot Properties}}   \\ \hline
\shortstack{Joint \\ Damping \\ (7DoF)}    & \shortstack{\\ {[}100, 100,\\ 100, 100, \\ 100, 10,\\ 0.4{]}} & \shortstack{\\ {[}2.0, 2.0, \\2.0, 2.0, \\ 2.0, 1.0, \\0.2{]}} & \shortstack{\\ {[}101.35, 100.06,\\ 100.08, 99.61, \\99.14,  10.05, \\0.83{]}} & \shortstack{\\ {[}1.06, 1.83,\\ 0.67, 0.41,\\  0.68, 0.64,\\ 0.57{]}} & \shortstack{\\ {[}98.11, 98.69,\\ 100.38, 99.64,\\ 101.01, 9.72,\\ 1.17{]}} & \shortstack{\\ {[}0.41,  0.63,\\  0.65, 0.15,\\ 0.96, 0.04,\\ 1.19{]}} & \shortstack{\\ {[}98.63, 99.28,\\ 95.83, 100.12,\\ 95.71,11.55,\\ 1.54 {]}} & \shortstack{\\ {[}1.23, 0.71,\\ 1.42, 0.81,\\ 0.77, 0.46,\\ 0.67{]}} \\ \hline
\multicolumn{9}{|c|}{\textbf{Gripper Properties (Left Right)}}   \\ \hline
  Sliding Friction  
  & {[}0.5, 0.5{]}     & {[}0.25, 0.25{]}       & {[}0.47, 1.78{]}    & {[}0.70, 0.86{]}    & {[}1.37, 0.76 {]}    & {[}0.35, 0.39 {]}  & {[}3.04, 2.68{]}   & {[}0.46, 0.54{]}                                                      \\ \hline
  Torsional Friction
   & {[}0.5, 0.5{]}    & ${[}0.25, 0.25{]}$     & {[}1.38, 1.84{]}   & {[}0.66, 0.92{]}    & {[}0.79, 2.29{]}   & {[}0.42, 0.80{]} & {[}1.04, 2.0{]}         & {[}0.72, 0.83{]}         \\ \hline
\end{tabular}
\vspace{-3mm}
\end{table*}}

\subsection{Sim-to-real transfer with optimized DR}
\label{sec:sec:DR}
% overview
While in many prior works, domain randomization methods have shown to be effective in addressing the reality gap, selecting the randomization ranges still remain challenging, and training the RL agent on overestimated ranges may be inevitable and could lead to poor learning performance. In the last experiment, we have successfully identified the parameter distributions, but we have yet demonstrated it to be effective to learn adequate policy and transfer to the real world. Therefore, in this experiment, we aimed at comparing the learning performance of policies learned from three methods, namely normal DR, DROID without DR and DROID with DR. These policies were trained to open the door without springs in the simulation and then directly transferred to the real world to open the same door. Normal DR approach was trained on the estimated parameter distributions which correspond to the left most columns in Tab.~\ref{tab:init_final_param}. DROID without DR, similar to SI, was trained using only fixed parameters. In this case, the parameters used were the optimized $\boldsymbol{\mu}$ listed in column four of the Tab.~\ref{tab:init_final_param}. Finally, DROID with DR was trained based on the optimized distributions, i.e. column four and five of Tab.~\ref{tab:init_final_param}.

For the sake of fairness, all of them were trained using model-free on-policy approach, \emph{i.e.} PPO, and shared identical RL hyperparameters, architectures of networks, observations, and reward functions, with the parameter ranges being the only differences. In the reinforcement learning setting, the observation included the joint positions and velocities of the arm, the relative positions between the robot gripper and the knob. These combined to form the 23 DoF state space. The action space consisted of 9 DoF, specifying the 7 robot joint positions and 2 gripper widths. The critic and actor models were represented by a neural network with two hidden layers of size 64, following tanh activation functions. We performed our simulations on MuJoCo physics engine with timestep of 0.001s. During the training, each episode took a maximum of 512 steps. We employed the ADAM optimizer with a stepsize of 0.001 and mini-batch of 64 episodes to update the policy and value networks. 

\begin{figure}[]
    \centering
    {\includegraphics[width=0.41\textwidth]{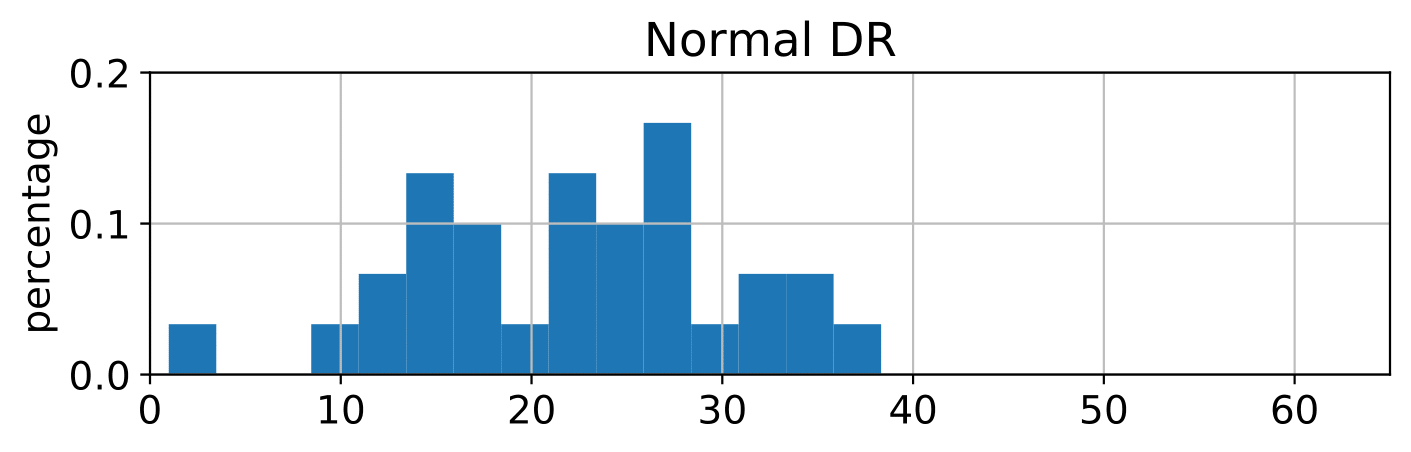} \\
    \includegraphics[width=0.41\textwidth]{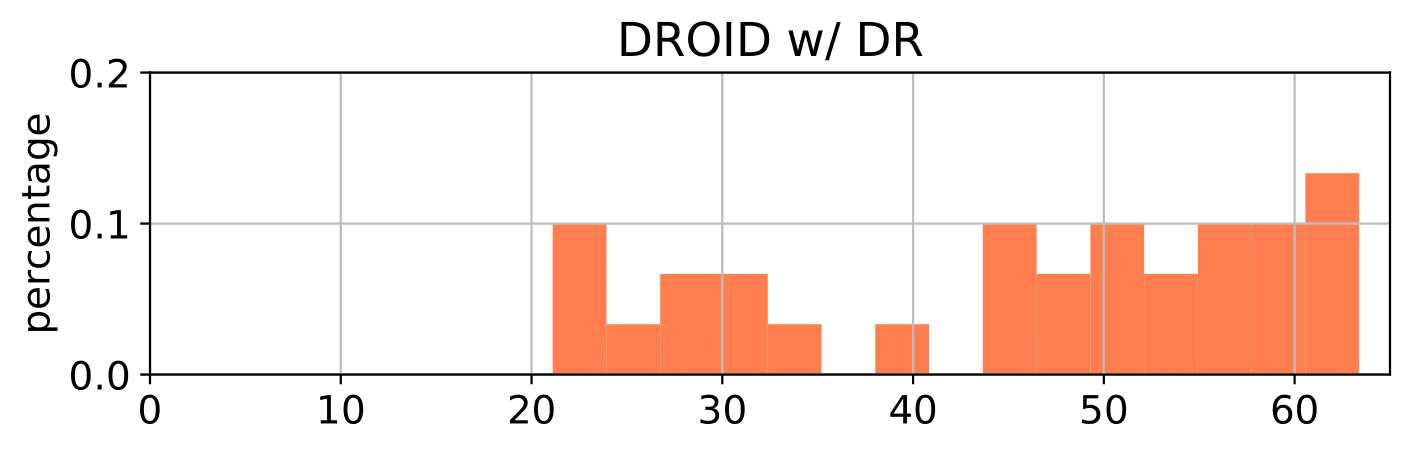} \\
    \includegraphics[width=0.41\textwidth]{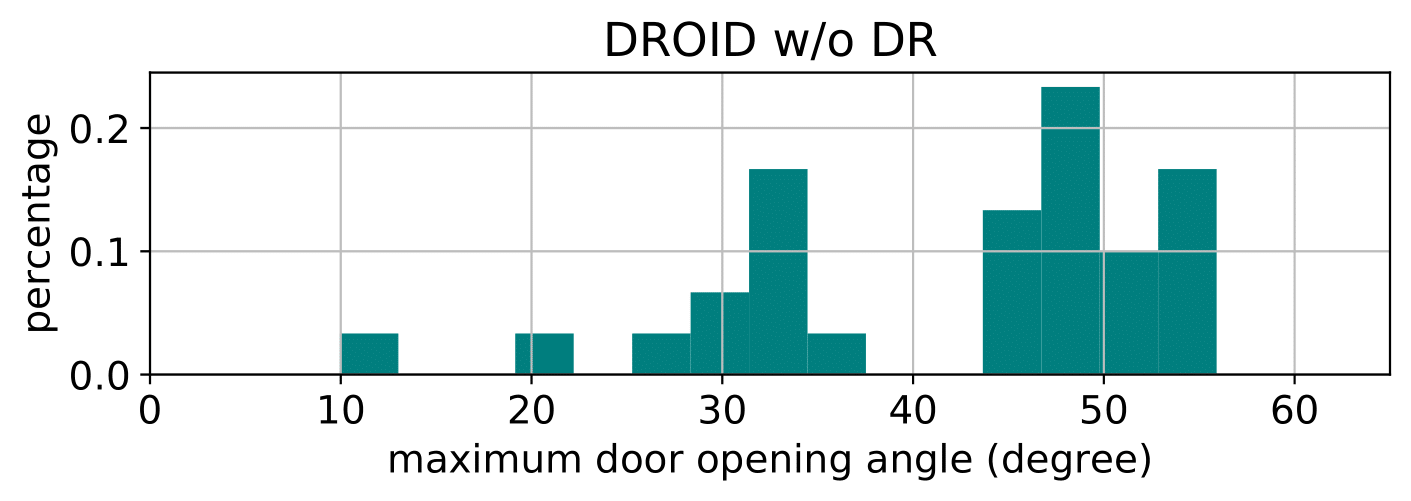}
    }
    \caption{Comparison of the door opening performance in reality for three methods (from top to bottom): normal DR, DROID with DR and DROID without DR. The horizontal axis is the maximal opening angle of the door in degrees. The vertical axis is the percentile value for each bin. It compares the distributions of maximal angles over 30 runs for each method. }
\label{fig:dr_comp}
\vspace{-2mm}
\end{figure}

We have trained three policies with each approach, and have tested each of them in the real world for 10 times.
The results for sim-to-real comparisons are displayed in Tab.~\ref{tab:sim2real_comp}. The door opening angles have been recorded each time and the historgram is shown in Fig.~\ref{fig:dr_comp}. We define a successful door opening case as the one with the hinge angle of the door larger than $30^\circ$ in real-world trials. From these results, we can see the later two approaches (DROID with DR and without DR) outperform the standard DR. In the DROID with DR setting, the optimized means and variances for DR ($80\%$ success rate) can achieve significant advantages in real-world tests over the normal DR ($20\%$ success rate) with initialized means and variances. We also notice that the policies trained without DR using optimized means $\mu_{opt}$ as system parameters in simulation can achieve the highest success rate of $86.7\%$ in reality, even higher than the DROID with DR method.
On one hand, this implies that the optimized $\mu_{opt}$ indeed achieves a good sim-to-real transfer. On the other hand, the fact that DROID without DR produces a higher success rate but lower door opening angles than DROID with DR is interesting. 
We attribute this to the relatively weak dependency on system dynamics when the door was opened with small angles (\emph{e.g.}, $30^\circ$), but as the hinge angle increased, the systems dynamics became more important for the door opening process due to the larger contact forces between the gripper and the door knob. Hence, the policies cannot further open the door well without a proper DR in simulated training process, which was also testified in Fig.~\ref{fig:dr_comp}. 
From the distributions of maximal door-opening angle, we show that DROID with DR can achieve the door opening with more concentration on larger maximal hinge angles, \emph{e.g.} around $60^\circ$, compared against normal DR and DROID without DR. Tab.~\ref{tab:sim2real_comp} also demonstrates that DROID with DR can achieve the highest average door-opening angle among all three methods, even though it does not perform as great as the normal DR method in simulation. As for the number of steps taken to open the door, our method have advantageous performances consistently in both simulation and reality, which indicates a faster door-opening process. 

% sim2real result comparisons
{\renewcommand{\arraystretch}{1.75}
\begin{table}[]
\centering
\captionof{table}{Comparisons of Sim-To-Real results}
\label{tab:sim2real_comp}
\begin{tabular}{|p{1.4cm}|c|c|c|c|c|} \hline
& \multicolumn{1}{c|}{}
& \multicolumn{1}{c|}{\shortstack{DR\\($\mu_{init}$, $\Sigma_{init}$)}} 
& \multicolumn{1}{c|}{$\mu_{opt}$} 
& \multicolumn{1}{c|}{\shortstack{DR\\($\mu_{opt}$, $\Sigma_{opt}$)}} \\ \hline
\multirow{ 2}{*}{\shortstack{\\ Success Rate\\(angle\textgreater30$^{\circ}$)}} & sim &  $100\%$      &   $100\%$  &   $100\%$        \\ \cline{2-5}
& real &   $20\%$     &   $\mathbf{86.7\%}$     &  $80\%$         \\ \hline
\multirow{ 2}{*}{\shortstack{\\ Open Angle \\ (mean$\pm$std)}} & sim &  $\mathbf{91.3}\pm0.4$      &  $70.8\pm16.0$   &   $91.2\pm0.2$       \\ \cline{2-5}
& real &  $22.4\pm8.2$      &     $42.2\pm11.1$     &    $\mathbf{45.4}\pm 13.6$       \\ \hline
   \multirow{ 2}{*}{\shortstack{\\ Open Steps\\ (mean$\pm$std)}} & sim &  $96.3\pm107.2$      &  $60.3\pm2.6$     &    $\mathbf{38.7}\pm14.1$      \\ \cline{2-5}
  & real &  $68.7\pm8.4$      &  $73.6\pm7.0$      &    $\mathbf{68.3}\pm 11.0$       \\ \hline
\end{tabular}
\vspace{-2mm}
\end{table}}

\subsection{Generalization of the Learned Policy}
As mentioned in the previous section, we only used one human demonstration to determine the parameter distribution. This demonstration was, however, not used during RL because it can only serve for a door with the same dimensions. In this experiment, we tested the policy with three additional door knob locations in the simulation and in the real system as shown in the Fig~\ref{fig:setup}(c) to emulate doors with different dimensions. These locations were set to be 5\emph{cm} apart along the lever arm of the door. With DROID, the trained policy was able to pull the door knob, at different locations, and open the door successfully. As such, we have verified that once the parameter distribution is identified using a human demonstration, different RL policies can be trained within the optimized distribution to accomplish tasks other than the one demonstrated. 

\section{CONCLUSION}
In this paper, we proposed a novel and generic framework, Domain Randomization Optimization IDentification (DROID) to minimize the reality gap between simulated and real environments. The approach is designed to be applicable to a range of contact-rich manipulation tasks, such as door opening. By executing a human demonstration trajectory in both simulation and reality, the differences in their dynamics can be minimized by iteratively updating and identify the distributions of the real dynamics parameters. Using a door opening task as an example, we have verified its capability to identify reasonable parameter distributions and thus reduce the reality gap. A successful RL policy can then be obtained by training in this distribution and directly transferring to the real world. The sim-to-real performance has shown to be superior than training with typical DR and SI approaches. Finally, we also demonstrated that a generalized RL policy can be trained to accomplish different tasks, given that the system dynamics remains unchanged.
%%%%%%%%%%%%%%%%%%%%%%%%%%%%%%%%%%%%%%%%%%%%%%%%%%%%%%%%%%%%%%%%%%%%%%%%%%%%%%%%

\addtolength{\textheight}{-12cm}   % This command serves to balance the column lengths

%%%%%%%%%%%%%%%%%%%%%%%%%%%%%%%%%%%%%%%%%%%%%%%%%%%%%%%%%%%%%%%%%%%%%%%%%%%%%%%%

\bibliographystyle{IEEEtran}
\bibliography{ICRA21}

\end{document}